\documentclass[conference]{IEEEtran}
\IEEEoverridecommandlockouts
\usepackage{graphicx}
\usepackage{float}
\usepackage{multirow}
\usepackage{array}
\usepackage{subcaption}
\usepackage{tabularx}
\usepackage{bm}
\usepackage[export]{adjustbox}
\graphicspath{ {images/} }
\usepackage{amsmath}
\usepackage{amssymb}
\usepackage{cite}
\usepackage{amsmath,amssymb,amsfonts}
\usepackage{algorithmic}
\usepackage{graphicx}
\usepackage{textcomp}
\usepackage{xcolor}

\def\BibTeX{{\rm B\kern-.05em{\sc i\kern-.025em b}\kern-.08em
    T\kern-.1667em\lower.7ex\hbox{E}\kern-.125emX}}
\begin{document}

\title{Quantifying error contributions of computational steps, algorithms and hyperparameter choices in image classification pipelines}

\author{Aritra~Chowdhury,
        Malik~Magdon-Ismail, ~\IEEEmembership{Member,~IEEE}
        and~B{\"u}lent~Yener,~\IEEEmembership{Fellow,~IEEE}

}


\maketitle

\begin{abstract}
Data science relies on pipelines that are organized in the form of interdependent computational steps. Each step consists of various candidate algorithms that maybe used for performing a particular function. Each algorithm consists of  several hyperparameters. Algorithms and hyperparameters must be optimized as a whole to produce the best performance. Typical machine learning pipelines  typically consist of complex algorithms in each of the steps. Not only is the selection process combinatorial, but it is also important to interpret and understand the pipelines. We propose a method to quantify the importance of different layers in the pipeline, by computing an error contribution relative to an \textit{agnostic} choice of algorithms in that layer. We demonstrate our methodology on image classification pipelines. The \textit{agnostic} methodology quantifies the error contributions from the computational steps, algorithms and hyperparameters in the image classification pipeline. We show that algorithm selection and hyper-parameter optimization methods can be used to quantify the error contribution and that random search is able to quantify the contribution more accurately than Bayesian optimization. This methodology can be used by domain experts to understand machine learning and data analysis pipelines in terms of their individual components, which can help in prioritizing different components of the pipeline.

\end{abstract}




\section{Introduction} 
\label{sec1}
Machine learning and data science have entered many domains of human effort in modern times. The number of self-reported data scientists has doubled in recent years \cite{harrison1995validity}. They have entered various domains including academia, industry and business among others. There has therefore been a demand for machine learning tools that are flexible, powerful and most importantly, interpretable. The effective application of machine learning tools unfortunately requires an expert understanding of the frameworks and algorithms that are present in a machine learning pipeline. It also requires knowledge of the problem domain and understanding of the assumptions used in the analysis. In order for tools to be used adequately by non-experts; new tools must be developed for understanding and interpreting the results of a data analysis pipeline in a specific domain.
\begin{figure}[ht!]
    \centering
    \includegraphics[width=0.5\textwidth]{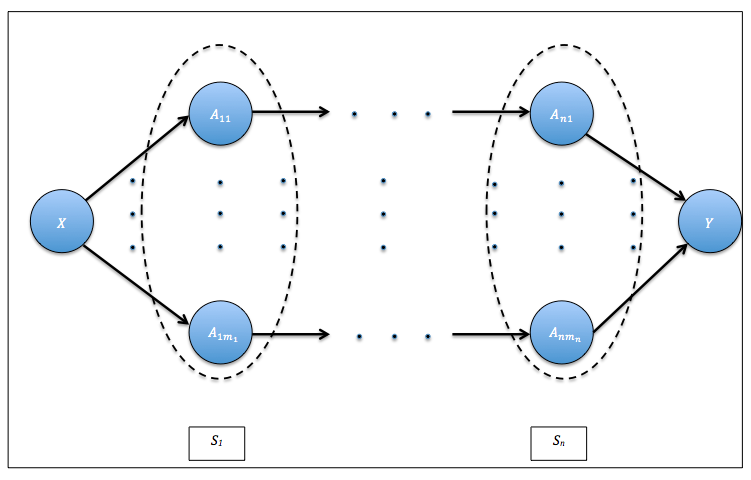}
    \caption{Representation of a data analysis pipeline. This is represented as a generalized directed acyclic graph. $S_i$ represents the $i$-th computational step in the pipeline and $A_{ij}$ represents the $j$-th algorithm in the $i$-th step. $X$ is the input dataset and $Y$ is the evaluation metric.}
    \label{fig:pipeline}
\end{figure}
Pipelines in machine learning and data science are commonly organized in the form of interdependent components. Such components that make up a data analysis pipeline include data preprocessing, feature extraction, feature transformation, model building and model evaluation among others. 
Fig. \ref{fig:pipeline} shows a generic representation of data analysis pipelines as a feed-forward network. Each computational step of the pipeline $S_{i}$ consists of several algorithms ($A_{ij}$) to choose from. Each algorithm in the pipeline has its own hyperparameters $\bm{\theta_{ij}}$ that must be optimized. Therefore, there are an exponential number of combinations of algorithms and hyperparameters in a given data analysis pipeline, which makes it computationally intensive task to optimize the pipeline. Tuning this pipeline can be viewed as the optimization of an objective function that is noisy and expensive to evaluate. The input to the pipeline is a dataset $X$, the pipeline $P$ (a network consisting of the steps $S_{i}$, the algorithms $A_{ij}$ and corresponding hyperparameters $\bm{\theta_{ij}}$) and the objective to optimize $Y$ such as validation error, accuracy, F1-score, or cross-entropy loss etc.The goal of a data scientist is to find the best set of algorithms and hyperparameters in this pipeline that optimizes the objective function. This corresponds to finding an optimal path through the pipeline in Fig. \ref{fig:pipeline}.  Simple methods such as grid and random search \cite{bergstra2012random} have been used to tackle this problem. More complicated approaches such as Bayesian optimization \cite{snoek2012practical, zhang2016flash} have been used successfully for approaching more difficult problems. Pipeline optimization as a whole has also been approached using genetic algorithms \cite{olson2016evaluation, olson2016tpot}.
We use grid search, random search and Bayesian optimization methods for optimization of the pipeline and each individual path in it. Our present goal is not to improve ways to optimize the pipeline, but to use any one such method to help a domain scientist quantify the importance of different steps in the pipeline. For example "How important is feature extraction?".

Domain experts prefer to understand how predictive decisions are made by the pipeline. Recently there has been an advent of models and techniques for improving the interpretability of machine learning. \cite{ribeiro2016model} introduces a model-agnostic method for interpreting the results of complex machine learning algorithms. 
\cite{koh2017understanding} uses influence functions to understand blackbox predictions. In this work, we attempt to provide an interpretation of machine learning pipelines in terms of the importance and sensitivity of components in the pipeline (steps, algorithms and hyperparameters) as opposed to the approaches which are geared toward interpretation of algorithms based on the dataset (see \cite{koh2017understanding}).Using our approach, one can understand the importance of different steps like feature extraction and feature transformation and individual algorithms and hyperparameters. To our knowledge, this type of approach to interpretation has not been taken before.
To this end, we propose the understanding of the contribution of error in data analysis pipelines using a method that we denote as the \textit{agnostic} methodology. Essentially, to quantify the contribution of a particular component, we compute the error from the pipeline when the component is selected \textit{agnostically}. We use the cross-entropy loss as the performance metric of the optimization algorithms and basis of error quantification in the image classification pipelines. 
Understanding the importance of the components in the predictive model is important for experts to design better data analysis pipelines. 
Experts can use the information from error contributions to focus attention on certain parts of the pipeline depending on the source of error. 
We introduce a methodology to quantify the contribution of error from different components of the data analysis pipeline, namely the computational steps and algorithms in the pipeline. 

Pipeline optimization methods and algorithms like grid search, random search \cite{bergstra2012random} and Bayesian optimization \cite{snoek2012practical} are used to optimize the pipeline for performing experiments with our \textit{agnostic} error contribution methodology. We take two different approaches to optimization. The first is hyper-parameter optimization (HPO) where a computational path in Fig. \ref{fig:pipeline} is optimized. The second type of optimization is denoted as combined algorithm selection and hyperparameter optimization (CASH). This term was introduced in \cite{thornton2013auto}. This is a more difficult problem, because the pipeline is optimized globally, in that the result of the optimization is a single optimized path that produces the best performance over all the paths in the machine learning workflow.  

We use four datasets to demonstrate the error contribution methodology. The problem we focus on is image classification. We show the performance of the optimization frameworks (HPO and CASH) for the experiments. We show experimentally that CASH using random search and Bayesian optimization can be efficiently used for quantification of errors from the different computational steps of the pipeline. In addition, HPO frameworks of both Bayesian optimization and random search provides  estimates of error contributions from the algorithms and hyperparameters in a particular path of the pipeline.
We demonstrate from the results that the \textit{agnostic} error contribution methodology maybe used by both data science and domain experts to improve and interpret the results of image classification pipelines. In addition, we observe that random search is a more accurate estimator of error contribution than Bayesian optimization. Finally, we demonstrate a visualization on a real pipeline.


\section{Foundations}
\label{sec2}
In this section we describe the optimization problem and methods that are used in this work. 
\subsection{Algorithm selection and hyper-parameter optimization}
\label{subsec_AS_HPO}
We approach the problem of optimization of the pipeline from two frameworks. In one framework, each path in the pipeline in Fig. \ref{fig:pipeline} is individually optimized. This essentially boils down to the problem of hyper-parameter optimization (HPO)  because the hyperparameters of each algorithm are optimized for each individual path. In the second framework, the entire pipeline is optimized. This means that the algorithms and hyperparameters are optimized together. This is denoted as combined algorithm selection and hyper-parameter optimization (CASH).

\subsubsection{Hyper-parameter optimization (HPO)}
\label{subsubsec_HPO}
Let the \textit{n} hyperparameters in a path be denoted as $\theta_1, \theta_2, ..., \theta_n$, and let $\Theta_1, \Theta_2, ..., \Theta_n$ be their respective domains. The hyperparameter space of the path is \textbf{$\Theta$} = $\Theta_1 \times \Theta_2 \times ... \times \Theta_n$.

When trained with $\emph{$\theta$} \in \textbf{$\Theta$}$ on data $D_{train}$, the validation error is denoted as \par
\begin{figure}[ht!]
\begin{adjustbox}{right}
   \begin{subfigure}{\columnwidth}
      \includegraphics[width=\linewidth]{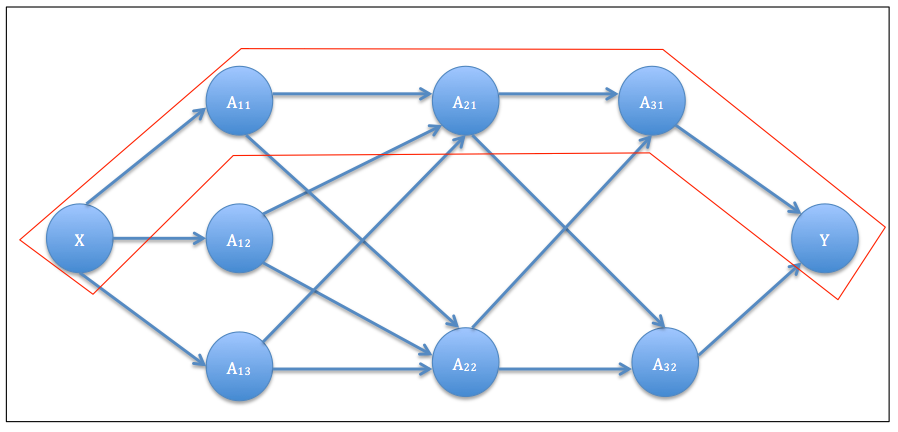}
      \caption{Hyper-parameter optimization in a data analysis pipeline. Each path in the pipeline is individually optimized.}
      \label{fig:HPO}
   \end{subfigure}
\end{adjustbox}

\begin{adjustbox}{right}
   \begin{subfigure}{\columnwidth}
      \includegraphics[width=\linewidth]{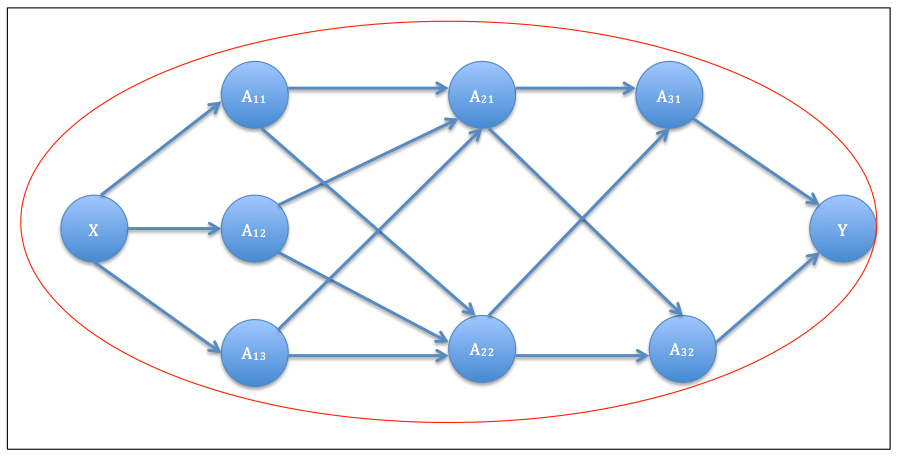}
      \caption{Combined algorithm selection and hyperparameter optimization (CASH) framework. The entire pipeline is optimized simultaneously.}
      \label{fig:CASH}
   \end{subfigure}
\end{adjustbox}
\caption{Optimization frameworks}\label{fig:frameworks}
\end{figure}

\noindent $\mathcal{L}(\theta, D_{train}, D_{valid})$. Using $k$-fold cross-validation, the hyperparameter optimization problem for a dataset $D$ is to minimize:
\begin{equation}
f^D(\theta) = \frac{1}{k}\sum_{i=1}^{k} \mathcal{L}(\emph{$\theta$}, D_{train}^{(i)}, D_{valid}^{(i)})
\label{eq:hpo}
\end{equation}
Hyperparameters $\theta$ may be numerical, categorical or conditional with a finite domain. The minimization of this objective function provides the optimal configuration of hyperparameters on a particular path in the pipeline in Fig. \ref{fig:pipeline}. The optimization of the objective function defined by Eq. \ref{eq:hpo} is very expensive. Depending on the type of hyper-parameter variables, the derivatives and convexity properties maybe unknown, and derivative free global optimization methods like Bayesian optimization and techniques like random search maybe used to tackle this problem. This framework is represented in Fig. \ref{fig:HPO}.

\subsubsection{Combined algorithm selection and hyper-parameter optimization (CASH)}
\label{subsubsec_CASH}
We can define the CASH formulation using Fig. 1. Let there be $n$ computational steps in the pipeline. Each step $i$ in the pipeline consists of algorithms $A_i(\Theta_i)$, where $A_i(\Theta_i) = \{A_{i1}(\theta_{i1}), ..., A_{im_{i}}(\theta_{im_{i}})\}$, $m_{i}$ is the number of algorithms in step $i$, $A_{ij}$ represents the $j$-th algorithm in step $i$, and \textbf{$\theta_{ij}$} represents the set of hyperparameters corresponding to  $A_{ij}$. The entire space of algorithms and hyperparameters is therefore given by \par
\noindent $\mathcal{A} = A_1(\Theta_1) \times A_2(\Theta_2) \times ... \times A_n(\Theta_n)$. The objective function to be minimized for CASH is given by
\begin{equation}
f^D(A) = \frac{1}{k}\sum_{i=1}^{k} \mathcal{L}(\emph{A}, D_{train}^{(i)}, D_{valid}^{(i)})
\label{eq:cash}
\end{equation}
where, $A \in \mathcal{A}$ and other notations are the same as those introduced in the previous section.
Similar to the objective function defined over the hyperparameters in Eq. \ref{eq:hpo}, the optimization in Eq. \ref{eq:cash} is even more difficult due to the additional problem of algorithm selection. Again, the derivates may be impossible to compute and convexity properties may be completely unknown. This framework is represented in Fig. \ref{fig:CASH}.


\subsection{Optimization methods}
\label{optimization}
The critical step in HPO or CASH is to choose the set of trials in the search space, which is $\Theta$ for HPO and $\mathcal{A}$ for CASH. In this section, methods that are used in this paper for optimization of Eq. \ref{eq:hpo} and Eq. \ref{eq:cash} are described. Grid search, random search and Bayesian optimization are used in this work.
\subsubsection{Grid search}
\label{grid}
Grid search is the simplest of all methods for coming up with trials in the search space. The set of trials in grid search is formed by assembling every possible set of values in $\Theta$ (HPO) and $\mathcal{A}$ (CASH) and computing the validation loss for each. The configuration $\theta \in \Theta$ or $A \in \mathcal{A}$ that minimizes the validation loss $\mathcal{L}$ is chosen as the optimum configuration. Unfortunately grid search is computationally very expensive. For HPO, the number of trials corresponds to $\prod_{i=1}^n |\Theta_i|$, and for CASH this is $\prod_{i=1}^n |A_i(\Theta_i)|$. This product makes grid search suffer from the \textit{curse of dimensionality}. This is because the number of trials grows exponentially with the number of hyperparameters. However, grid search has certain advantages. Firstly, parallelization and implementation is trivial. In addition, grid search is robust in the sense that results maybe replicated easily. 

\subsubsection{Random search}
\label{random}
Random search is the optimization method where trial configurations are randomly sampled from the search space of $\Theta$ (HPO) or $\mathcal{A}$ (CASH). \cite{bergstra2012random} shows empirically and theoretically that randomly selecting trials is sufficiently accurate and more efficient than performing optimization using grid search. We also show similar results in this work.

\subsubsection{Bayesian optimization}
\label{Bayesian}
Sequential model based Bayesian optimization (SMBO) \cite{hutter2011sequential} is the method of choice when it comes to optimization of complicated black-box functions. In a nutshell, it consists of two components. The first is a probabilistic model and the second is an acquisition function. The probabilistic model can be modelled using Gaussian processes (Spearmint) \cite{snoek2012practical}, random forests (SMAC) \cite{hutter2011sequential} and using density estimation with Tree-structured Parzen estimators (TPE) \cite{bergstra2011algorithms}. The acquisition function determines the future candidates or trials for evaluation. The acquisition function is relatively cheap to evaluate compared to the actual objective function $f^D$. One of the most prominent acquisition functions is \textit{expected improvement} (EI) \cite{expected_improvement}. We use the sequential model-based algorithm configuration (SMAC) that uses random forests as the Bayesian optimization framework. This is because it can be used for optimizing conditional hyperparameter configurations. The choice is also based on empirical results in \cite{eggensperger2013towards}.

\section{Proposed methods}
\label{sec3}
In this section the proposed methodology for quantification of error contribution is presented. The method is independent of the optimization methods that maybe used for both the HPO and CASH formulations.

\subsection{Error contribution with the agnostic methodology}
\label{EQ}

We propose an \textit{agnostic} methodology for quantifying error contributions from different parts of the pipeline. It is defined as the minimum error obtained by being agnostic to a particular component of the pipeline (computational step, algorithms or hyperparameters). We shall define what \textit{agnostic} refers to for both computational steps, algorithms and hyperparameters individually.

\subsubsection{Quantification of error from computational steps}
\label{subsubsec_eq_steps}
The \textit{agnostic} methodology maybe used for quantification of contributions from computational steps like feature extraction, data pre-processing and learning algorithms. Being \textit{agnostic} to a computational step means that the algorithms in that step are selected randomly for that step while the remaining pipeline is optimized. The average of the minimum errors obtained with each algorithm in the step used as the only algorithm in that particular step, provides an estimate of the agnostic error from a particular pipeline.  
More formally, the agnostic methodology is implemented for computational steps in the following manner. Using Fig. \ref{fig:pipeline} as a reference, let $n$ be the number of steps in the pipeline. Each step in the pipeline is denoted as $S_i$. $|S_i|$ is the number of algorithms in step $i$. $A_{ij}$ denotes the $j$-th algorithm in the $i$-th step. $E^*$ represents the minimum validation error found after optimization of the entire pipeline (using the CASH framework). $E_{A_{ij}}^*$ is the minimum  validation error found with $A_{ij}$ as the only algorithm in step $i$. The error contribution from step $i$, $EC_{S_i}^*$ is given by Eq. \ref{eq_step}.
\begin{equation}
\label{eq_step}
EC_{S_i}^* = \frac{1}{|S_i|}\sum_{z=1}^{|S_i|} E_{A_{iz}}^* - E^*,
\end{equation}
where, $i = {1, ..., n}, j = {1, ..., |S_i|}$
Taking the difference with respect to the global minimum in Eq. \ref{eq_step} provides an estimate of the error contribution from step $i$ of the pipeline. A large value of $EC_{S_i}^*$ would mean that step $S_i$ is important for the pipeline.

\subsubsection{Quantification of error from algorithms}
\label{subsubsec_eq_alg}
The \textit{agnostic} methodology for algorithms is implemented as follows. Similar to the \textit{agnostic} methodology for steps, we define the \textit{agnostic} methodology for algorithms. In this case, we focus on a single path in the pipeline in Fig. 1. Let's assume we are trying to quantify the error contribution of a particular algorithm $A_{ij}$ that lies on path $p$. Being $agnostic$ to $A_{ij}$ means we optimize everything else on the path except the algorithm. This means that we pick the hyperparameters $\theta_{ij}$ of algorithm $A_{ij}$ randomly while optimizing the rest of the algorithms on the path. This is formally calculated by taking the average of the optimum errors on the path for each configuration of $\theta_{ij}$. The minimum validation error on the path is then subtracted from this error to give us the error contribution from algorithm $A_{ij}$ on path $p$. These errors are computed using the results and the search trials on the CASH framework in section \ref{subsubsec_CASH}.

\begin{equation}
\label{eq_alg}
EC_{A_{ij}}^* = \frac{1}{|\theta_{ij}|}\sum_{z=1}^{|\theta_{ij}|} {E_{A_{ij}}^z}^* - {E_{A_{ij}^p}^*},
\end{equation}

where, $i = {1, ..., n}, j = {1, ..., |\theta_{ij}|}$, $|\theta_{ij}|$ represents the number of hyperparametric configurations of $A_{ij}$,  ${E_{A_{ij}}^z}^*$ is the minimum error obtained with the $z$-th configuration of $\theta_{ij}$ and $E_{A_{ij}^p}^*$ is the minimum error found over the path $p$ that consists of algorithm $A_{ij}$.

\subsubsection{Quantification of error from hyperparameters}
\label{subsubsec_eq_hyper}
The \textit{agnostic} methodology for hyperparameters is implemented as follows. In the case of hyperparameters, we focus on a single path similar to what we did for algorithms. Let's assume we are trying to quantify the error contribution of a particular hyper-parameter $\theta_{ijk}$ that lies on path $p$, i.e. the $k$-th hyper-parameter of the $j$-th algorithm in the $i$-th step of the pipeline. Being $agnostic$ to $\theta_{ijk}$ means we optimize everything else on the path except the hyper-parameter. This means that we pick the hyperparameter $\theta_{ijk}$ of algorithm $A_{ij}$ randomly while optimizing the rest of the hyperparameters on the path. This is formally calculated by taking the average of the optimum errors on the path for each configuration of $\theta_{ijk}$. The minimum validation error on the path is then subtracted from this error to give us the error contribution from hyper-parameter $\theta_{ijk}$ on path $p$. This is again computed using the HPO framework described in section \ref{subsubsec_HPO}.

\begin{equation}
\label{eq_hyper}
EC_{\theta_{ijk}}^* = \frac{1}{|\theta_{ijk}|}\sum_{z=1}^{|\theta_{ijk}|} {E_{\theta_{ijk}}^z}^* - {E_{A_{ij}^p}^*},
\end{equation}

where, $i = {1, ..., n}, j = {1, ..., |\theta_{ij}|}$, $k$ = number of hyperparameters of algorithm $A_{ij}$. $|\theta_{ijk}|$ represents the number of configurations of $\theta_{ijk}$,  ${E_{\theta_{ijk}}^z}^*$ is the minimum error obtained with the $z$-th configuration of $\theta_{ijk}$ and $E_{A_{ij}^p}^*$ is the minimum error found over the path $p$ that consists of algorithm $A_{ij}$.

\section{Experiments and results}
\label{sec4}

In this section, we describe the experiments performed on the data analysis pipeline to quantify the error contributions from different components of the pipeline. Image classification is the data analysis problem chosen for demonstrating the error quantification experiments. 

\begin{figure}[ht!]
    \centering
    \includegraphics[width=0.5\textwidth]{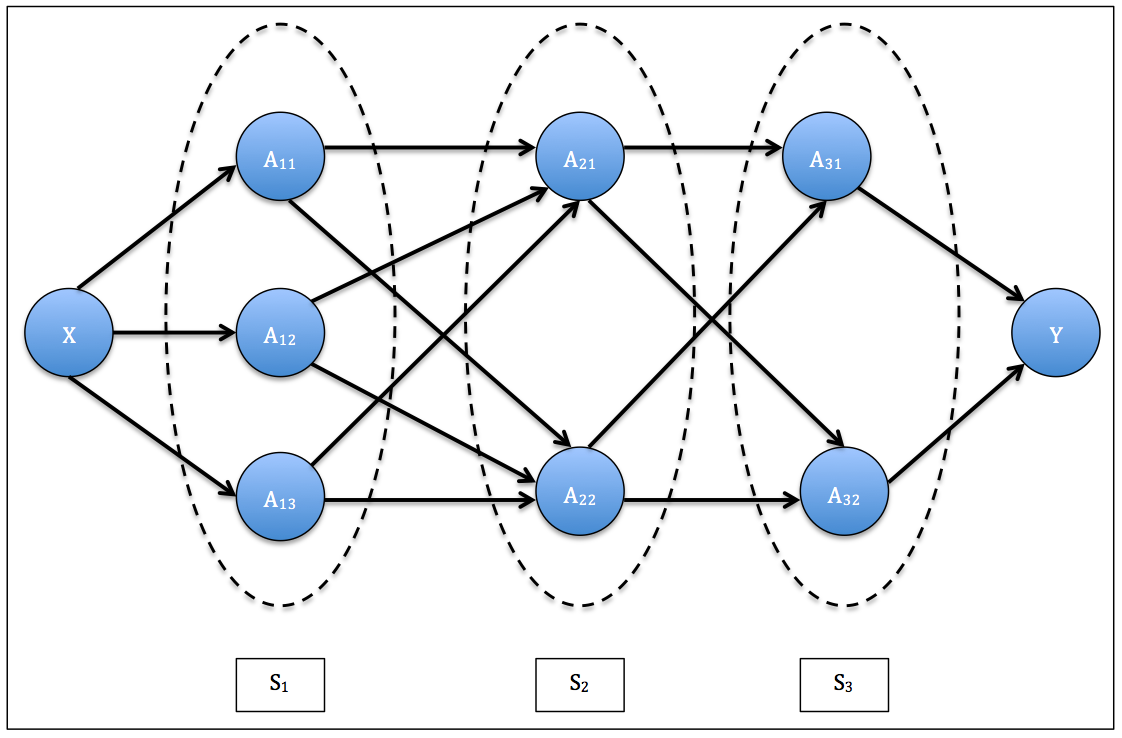}
    \caption{Representation of the image classification pipeline as a directed acyclic graph used in this work. This is an instantiation of the generalized data analysis pipeline in Fig. \ref{fig:pipeline}}
    \label{fig:images_pipeline}
\end{figure}

 The above figure shows the pipeline used in this work. There are 3 computational steps in this pipeline, namely feature extraction ($S_1$), feature transformation ($S_2$) and learning algorithms ($S_3$). The steps, algorithms and corresponding hyperparameters $A_{ij}(\theta_{ij})$ is described in Table \ref{table:algorithms_table}.

\begin{table}[ht!]
\centering
\caption{Algorithms and hyperparameters used in the image classification pipeline. The specific algorithms and corresponding \textit{hyperparameters} are defined in the last column}
\begin{tabularx}{\linewidth}{ |X|X|X| } 
 \hline
 Step & $A_{ij}(\theta_{ij})$ & Definition \\ 
 \hline
 \multirow{3}{*}{Feature extraction} & $A_{11}(\theta_{11})$ & Haralick texture features (\textit{Haralick distance}) \\ 
 & $A_{12}(\theta_{12})$ & Pre-trained CNN trained on ImageNet \cite{deng2009imagenet} database with VGG16 \cite{simonyan2014very} network  \\
  & $A_{13}(\theta_{13})$ & Pre-trained CNN trained on ImageNet \cite{deng2009imagenet} database with Inception \cite{szegedy2016rethinking} network  \\
 \hline
 \multirow{3}{*}{Feature transformation} & $A_{21}(\theta_{21})$ & PCA ($whitening$) \cite{wold1987principal} \\
 & $A_{22}(\theta_{22})$ & ISOMAP (\textit{number of neighbors, number of components}) \cite{tenenbaum2000global} \\
 \hline
 \multirow{3}{*}{Learning algorithms} & $A_{31}(\theta_{31})$ & Random forests (\textit{number of trees, maximum features}) \cite{breiman2001random} \\
 & $A_{32}(\theta_{32})$ & SVM ($C, \gamma$) \cite{cortes1995support}\\
 \hline
 \end{tabularx}
 \label{table:algorithms_table}
\end{table}
The algorithms defined in the table are selected for making up the components of the pipeline in Fig. \ref{fig:images_pipeline}. This is meant to serve as an example for demonstrating the experiments using the error contribution framework described in section \ref{sec3}. It can easily be generalized to any data analysis problem that involve pipelines.

\subsection{Optimization frameworks}
\label{frameworks}
Experiments are performed using two optimization frameworks. These frameworks have been described in detail in Section \ref{subsec_AS_HPO}. 
The first global optimization framework is the CASH framework described in Section \ref{subsubsec_CASH}. Here, the pipeline is optimized as a whole including the algorithms, which are themselves considered as hyperparameters in this framework.  Fig. \ref{fig:CASH} is a representation of this. This is used for quantification of the contribution of error with respect to \textit{computational steps} in the pipeline.

The second framework is the hyperparameter optimization (HPO) framework where each path in the pipeline is optimized individually. This is described in detail in section \ref{subsubsec_HPO}. This framework is used for quantifying the contribution of error with respect to algorithms and hyperparameters in the each path of the pipeline. The framework is depicted in Fig. \ref{fig:HPO}.

Specifically, we choose the path \textit{haralick texture features} - \textit{PCA} - \textit{random forests} to demonstrate the error quantification approach for algorithms and hyperparameters.

\begin{table}[ht!]
\centering
\caption{Description of the datasets used in this work}
\begin{tabularx}{\linewidth}{ |X|X|} 
 \hline
 Dataset (notation) & Distribution of classes \\ 
 \hline
 Breast cancer (\textit{breast}) \cite{bilgin2007cell} & \textit{benign}: 151, \textit{in-situ}: 93, \textit{invasive}: 202\\
 \hline
 Brain cancer (\textit{brain}) \cite{gunduz2004cell} & \textit{glioma}: 16, \textit{healthy}: 210, \textit{inflammation}: 107\\
 \hline
  Material science 1 (\textit{matsc1}) \cite{chowdhury2016image} & \textit{dendrites}: 441, \textit{non-dendrites}: 132 \\
 \hline
 Material science 2 (\textit{matsc2}) \cite{chowdhury2016image} & \textit{transverse}: 393, \textit{longitudinal}: 48 \\
 \hline
 \end{tabularx}
\label{table:datasets}
\end{table}

\subsection{Datasets}
Four datasets from the domains of medicine and material science are used in this work. They are image datasets of breast cancer\cite{bilgin2007cell}, brain cancer \cite{gunduz2004cell}, and two datasets of microstructures in material science \cite{chowdhury2016image}. They are described in Table \ref{table:datasets}.

\begin{table}[ht!]
\centering
\caption{Hyper-parameter domains corresponding the algorithms in Table \ref{table:algorithms_table} used by the optimization methods}
\begin{tabularx}{\linewidth}{ |X|X|X|X| } 
 \hline
 Algorithm & Hyper-parameter & Description & Values \\ 
 \hline
 Haralick texture feature & \textit{Haralick distance} & The Haralick distance to consider while computing the co-occurence matrix & [1, 2, 3, 4]\\
 \hline
 PCA & \textit{whitening} & Flag variable for whitening the data & [True, False]\\
 \hline
  ISOMAP & \textit{number of neighbors} & Number of neighbors to consider for each point & [3, 4, 5, 6, 7] \\
 \hline
 ISOMAP & \textit{number of components} & Number of co-ordinates for the manifold in ISOMAP algorithm & [2, 3, 4]\\
 \hline
 Random forests & \textit{number of estimators} & Number of trees in the forest & [8,  81, 154, 227, 300]\\
 \hline
 Random forests & \textit{maximum features} & The fraction of the total number of features to consider when looking for the best split & [0.3, 0.5, 0.7]\\
 \hline
 SVM & \textit{C} & Penalty parameter of the error term & [0.1, 25.075, 50.05, 75.025, 100.0] \\
 \hline
 SVM & $\gamma$ & Kernel coefficient for the radial basis function & [0.3, 0.5, 0.7] \\
 \hline
 \end{tabularx}
 
\label{table:hyper}
\end{table}

\begin{figure*}[ht!]
\centering
\begin{subfigure}{.5\textwidth}
  \centering
  \includegraphics[scale=0.5]{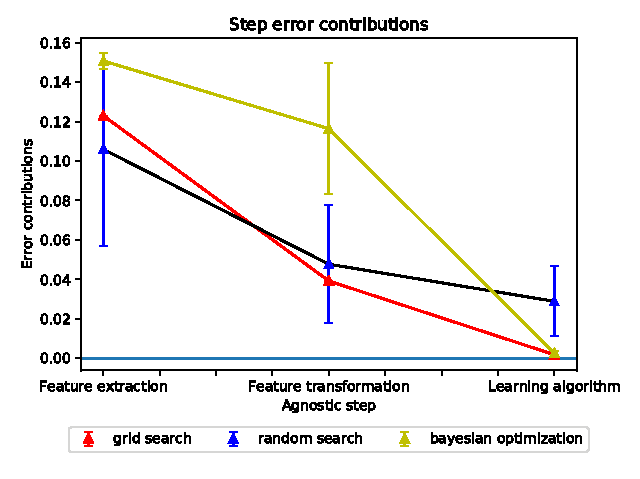}
  \caption{\textit{breast}}
  \label{fig:eq_steps_breast}
\end{subfigure}%
\begin{subfigure}{.5\textwidth}
  \centering
  \includegraphics[scale=0.5]{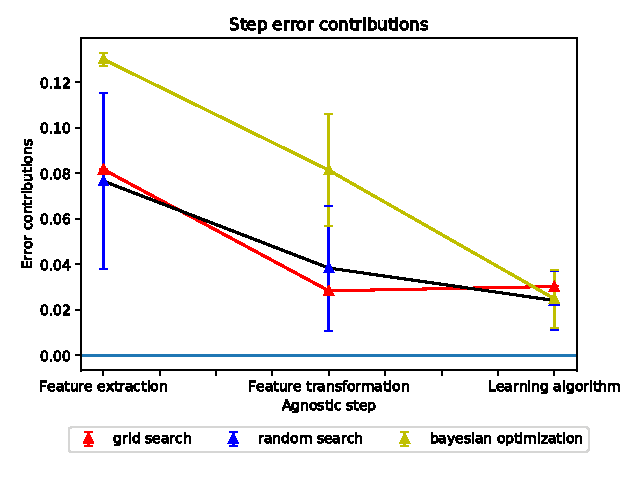}
  \caption{\textit{brain}}
  \label{fig:eq_step_brain}
\end{subfigure}
\begin{subfigure}{.5\textwidth}
  \centering
  \includegraphics[scale=0.5]{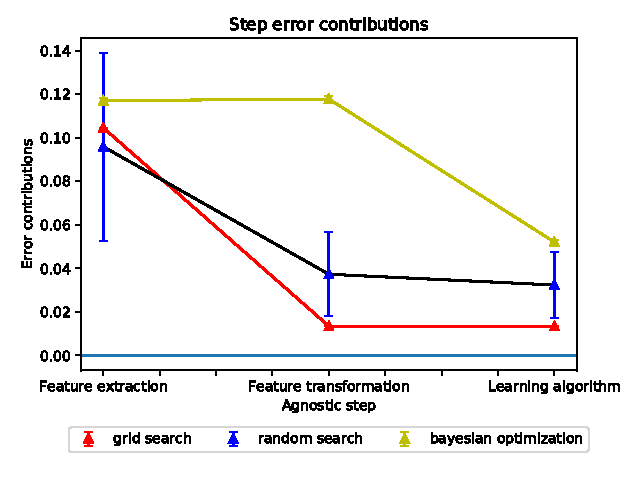}
  \caption{\textit{matsc1}}
  \label{fig:eq_steps_matsc1}
\end{subfigure}%
\begin{subfigure}{.5\textwidth}
  \centering
  \includegraphics[scale=0.5]{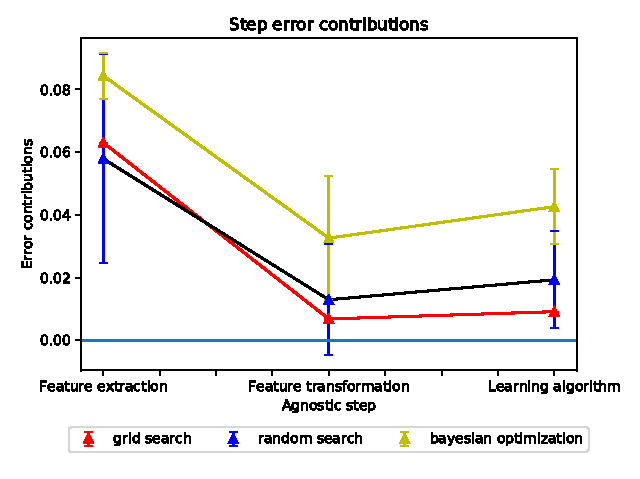}
  \caption{\textit{matsc2}}
  \label{fig:eq_steps_matsc2}
\end{subfigure}
\caption{Plots of error contributions from computational steps in the pipeline. The x-axis represents the steps in the pipeline - feature extraction, feature transformation and learning algorithms. The y-axis shows the values of the contributions from the corresponding steps in the pipeline. The maximum contribution in terms of error is from the feature extraction step in the pipeline. Random search (blue) follows the behavior of grid search (red) more accurately than Bayesian optimization (yellow). Hence, random search maybe used to quantify the error contributions instead of grid search.}
\label{fig:eq_steps}
\end{figure*} 

Table \ref{table:datasets} represents datasets from the scientific domain. These datasets have been chosen because they represent examples of real world datasets. They are noisy in the sense that they have artefacts in the images, are heavily imbalanced and are small in terms of number of samples. They are different from the very large datasets like ImageNet \cite{deng2009imagenet}, where deep learning techniques like convolutional neural networks have been shown to be superior. Even though deep neural networks represent an end-to-end workflow where the input image is fed into the network and the output classification is obtained at the other end, they may also be represented as pipelines, if the hyperparameters of the network are considered. \cite{shin2016deep} has shown that machine learning problems involving datasets from medical imaging may be solved using pre-trained and fine-tuned neural networks rather than training them from scratch. We have therefore used pre-trained models such as VGGnet \cite{simonyan2014very} and InceptionNet \cite{szegedy2016rethinking} as pre-trained feature extraction models that fit naturally in the pipeline framework described here for the purpose of illustrating the error quantification methodology.

\begin{figure*}[ht!]
\centering
\begin{subfigure}{.5\textwidth}
  \centering
  \includegraphics[scale=0.5]{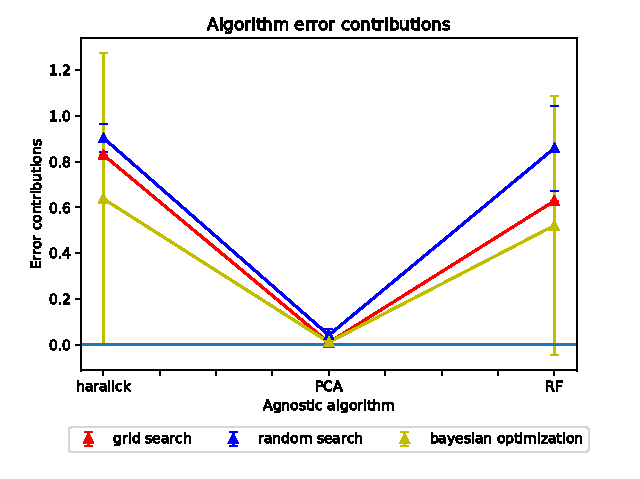}
  \caption{\textit{breast}}
  \label{fig:eq_alg_breast}
\end{subfigure}%
\begin{subfigure}{.5\textwidth}
  \centering
  \includegraphics[scale=0.5]{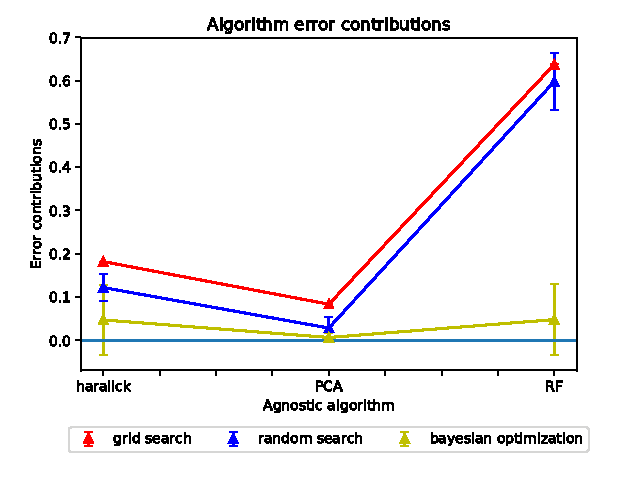}
  \caption{\textit{brain}}
  \label{fig:eq_alg_brain}
\end{subfigure}
\begin{subfigure}{.5\textwidth}
  \centering
  \includegraphics[scale=0.5]{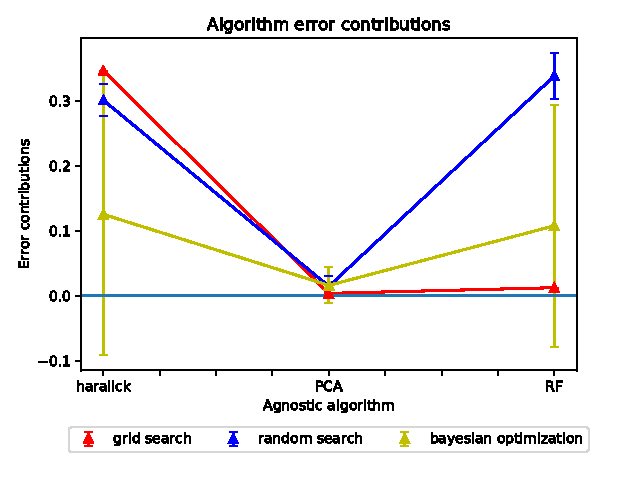}
  \caption{\textit{matsc1}}
  \label{fig:eq_alg_matsc1}
\end{subfigure}%
\begin{subfigure}{.5\textwidth}
  \centering
  \includegraphics[scale=0.5]{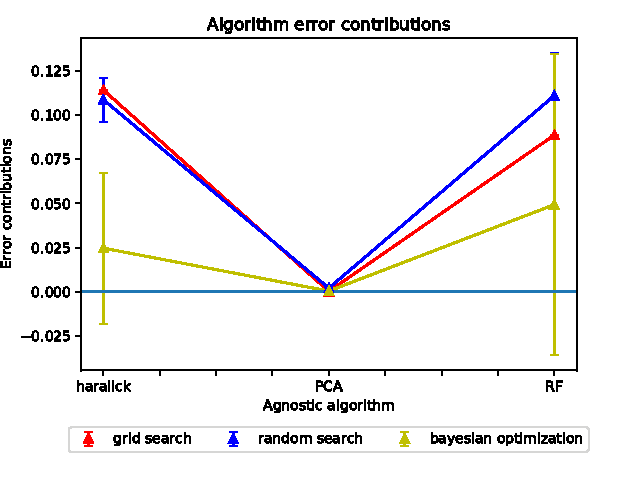}
  \caption{\textit{matsc2}}
  \label{fig:eq_alg_matsc2}
\end{subfigure}

\caption{Plots of error contributions from algorithms in the pipeline. The x-axis represents the algorithms in the path - \textit{haralick texture features}, $PCA$ and \textit{random forests}. The y-axis shows the values of the contributions from the corresponding algorithms in the path. Random search again mirrors the trend of grid search more than Bayesian optimization. Therefore random search maybe used instead of grid search for computing the contribution of error from algorithms in a path. The plots also show that it is more important to tune \textit{haralick texture features}, and \textit{random forests} than it is to tune  $PCA$. }
\label{fig:eq_alg}
\end{figure*}

\subsection{Error quantification experiments}
\label{eq_expts}
\begin{figure*}[ht!]
\centering
\begin{subfigure}{0.5\textwidth}
  \centering
  \includegraphics[scale=0.5]{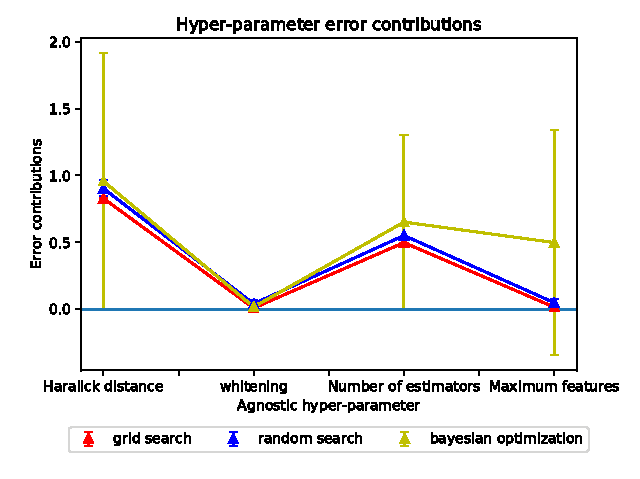}
  \caption{\textit{breast}}
  \label{fig:eq_hyper_breast}
\end{subfigure}%
\begin{subfigure}{.5\textwidth}
  \centering
  \includegraphics[scale=0.5]{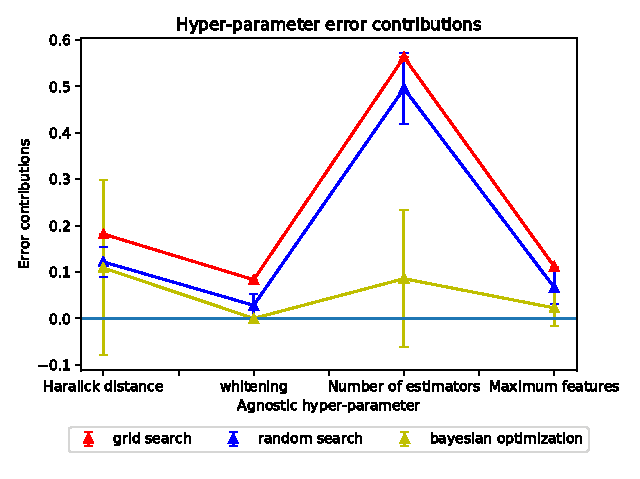}
  \caption{\textit{brain}}
  \label{fig:eq_hyper_brain}
\end{subfigure}
\begin{subfigure}{.5\textwidth}
  \centering
  \includegraphics[scale=0.5]{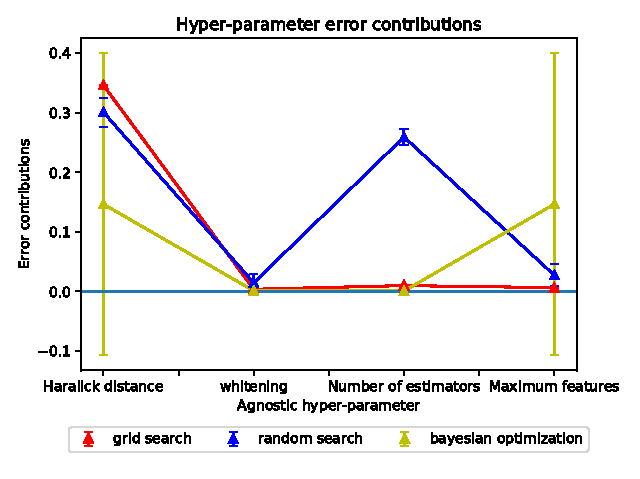}
  \caption{\textit{matsc1}}
  \label{fig:eq_hyper_matsc1}
\end{subfigure}%
\begin{subfigure}{.5\textwidth}
  \centering
  \includegraphics[scale=0.5]{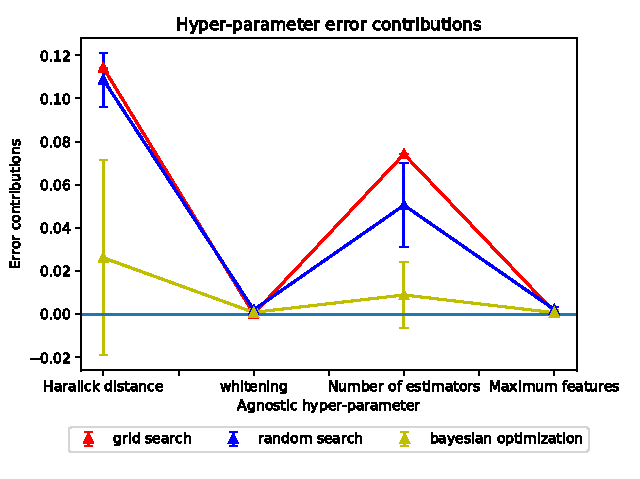}
  \caption{\textit{matsc2}}
  \label{fig:eq_hyper_matsc2}
\end{subfigure}

\caption{Plots of error contributions from hyperparameters in the pipeline. The x-axis represents the algorithms in the path - \textit{Haralick distance}, \textit{whitening}, \textit{Number of estimators} and \textit{Maximum features}. The y-axis shows the values of the contributions from the corresponding hyperparameters in the path. Random search again mirrors the trend of grid search more than Bayesian optimization. Therefore random search maybe used instead of grid search for computing the contribution of error from hyperparameters in a path. The plots also show that it is more important to tune the hyperparameters \textit{Haralick distance}, and \textit{Number of estimators} than it is to tune  the other hyperparameters.}
\label{fig:eq_hyper}
\end{figure*}

Experiments based on the quantification of error contributions framework described in section \ref{EQ} are presented here. The plots are of the error contribution values calculated using Eqs. \ref{eq_step}, \ref{eq_alg} and \ref{fig:eq_hyper} on the 4 datasets described in Table \ref{table:datasets}. 
\subsubsection{Experimental setting}
Optimization using the 3 algorithms in section \ref{optimization} is performed using the pipeline in Fig. \ref{fig:images_pipeline} on the 4 datasets in Table \ref{table:datasets}. The domain and possible values of the hyperparameters are described in Table \ref{table:hyper}. 
The convergence criteria (a hyper-hyper-parameter) is set at 50 iterations of unchanging  function value for each of the optimization methods. The choice of the convergence criteria and hyperparameters are independent of the error quantification methods. Results maybe obtained by using any choice of values for these components. The continuous hyperparameters \textit{maximum features}, \textit{C} and $\gamma$ have been discretized specifically for comparison of the optimization methods with grid search. In general, discretization of the hyperparameters is not necessary for performing optimization.

The error contribution values are obtained from the trials in the optimization methods described in Section \ref{sec2}. Grid search is only run once while the other algorithms are averaged over 5 runs with the mean and standard deviation shown in the following plots. These results are computed on the validation error (cross-entropy loss) obtained at the end of the pipeline. Random search and Bayesian optimization (using the SMAC algorithm) are implemented on both the frameworks described in Section \ref{frameworks}. The grid search results maybe used as the gold standard to compare the performance of other optimization algorithms. 

\subsubsection{Error contribution from computational steps}
The mean and standard deviation values of $EC_{S_i}$ calculated using Eq. \ref{eq_step} is represented in Fig. \ref{fig:eq_steps} for the 4 datasets in Table \ref{table:datasets}. The error is calculated using the formulation of $EC_{S_i}$ in section \ref{subsubsec_eq_steps}. We observe that most of the contribution comes from feature extraction algorithms. This means that it is most important to optimize the feature extraction step among the other steps as it is of most importance based on the plots. This confirms our intuitive belief that feature extraction algorithms are the most important components of a machine learning pipeline. 

Let us look at Fig. \ref{fig:eq_steps_breast} (contribution of the steps on the \textit{breast}) as an example. Here we see that the contribution from the steps reduces in magnitude as we move towards the end of the pipeline. This trend is quantified by all of the three methods - grid search, random search and Bayesian optimization. However, Bayesian optimization is not able to capture the contribution from feature transformation as accurately as random search with respect to grid search.
 The standard deviation of grid search is 0 because it was only run once for each dataset due to the time required for computation and also because of the robustness of the grid search method (the results don't change because we try out every single configuration). We observe from the results, that random search follows the behavior of grid search. It mirrors the behavior of grid search, but the results are not robust because of the high standard deviation. This is expected because the search trials may not include all the configurations in the pipeline as opposed to grid search where all the configurations of algorithms and hyperparameters are evaluated. The results of SMAC (Bayesian optimization) in the CASH framework do not follow the behavior of grid search as closely as random search. This is because the trials for SMAC are even more sparse with respect to the algorithms it selects for optimization of  the error in Eq. \ref{eq_step}. SMAC only samples a few configurations based on the updated probabilistic model as it narrows in on the optimum configuration. Therefore, it sometimes gives erroneous results as can be seen by the contribution of feature transformation in Fig. \ref{fig:eq_steps}.

\subsubsection{Error contribution from algorithms}

In Fig. \ref{fig:eq_alg}, the error contributions from algorithms are quantified using the formulation in Section \ref{subsubsec_eq_alg}. We select \textit{haralick texture features}, $PCA$ and \textit{random forests} as the path to demonstrate the contribution of error from algorithms, because each of these algorithms are associated with one or more hyperparameters. We observe a trend here, in that, the contribution from \textit{haralick texture features} and \textit{random forests} is more than $PCA$. This means that it is more important to tune \textit{haralick texture features}, and \textit{random forests} than it is to tune  $PCA$. This maybe attributed to the fact that the search space of hyperparameters (in Table \ref{table:hyper}) for \textit{haralick texture features} and \textit{random forests} is larger than that of \textit{PCA}. Again, we see the trend that random search performs better and is more robust in terms of following the behavior of grid search than Bayesian optimization.

Let us take the example of the error contribution of algorithms over the selected path in the \textit{brain} dataset depicted in Fig. \ref{fig:eq_alg_brain}. We observe that \textit{random forests} has the most amount of contribution with respect to the error followed by \textit{haralick texture features} and \textit{PCA} respectively. Again we see that Bayesian optimization is not able to capture the error contributions from \textit{random forests} adequately.

\begin{figure*}[ht!]
    \centering
    \includegraphics[width=\textwidth]{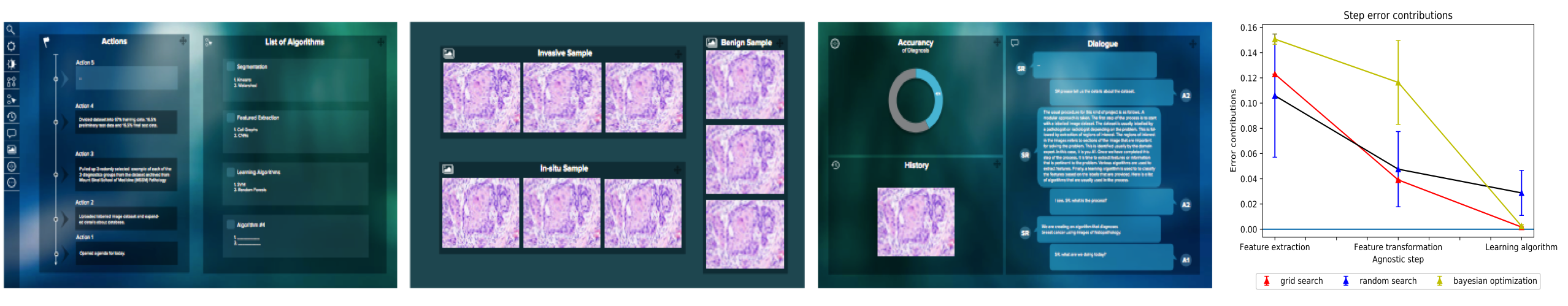}
    \caption{Front-end of the application that shows the interactive screens of the CIR. The first screen from the left shows the actions taken by the experts in the room, and the list of algorithms in the pipeline, The second screen shows a module where the experts may interact with the CIR. The third screen shows the dialogue among the experts and the results of the analysis. The right most screen shows how the error contribution results maybe used by domain experts and data scientists in the room to solve the problem of breast cancer diagnosis.}
    \label{fig:CISL}
\end{figure*}

\subsubsection{Error contribution from hyperparameters}

Fig. \ref{fig:eq_hyper} shows the error contributions from hyperparameters quantified using the formulation in Section \ref{subsubsec_eq_hyper}. We again select \textit{haralick texture features}, $PCA$ and \textit{random forests} as the path to demonstrate the contribution of error from algorithms. The hyperparameters (from Table \ref{table:hyper}) in this path are \textit{Haralick distance}, \textit{whitening}, \textit{Number of estimators} and \textit{Maximum features}. Again, we see the trend that random search performs better in terms of following the behavior of grid search than Bayesian optimization. In addition, we observe that it is in general more important to tune the hyperparameters \textit{Haralick distance}, and \textit{Number of estimators} than it is to tune  the other hyperparameters. This again could be due to the number of configurations of the respective hyperparameters in Table \ref{table:hyper}, where a hyper-parameter with a larger search space has more contribution to the error and is therefore more important to tune.

Let us look at the plot in Fig. \ref{fig:eq_hyper_brain} (contribution of hyperparameters with respect to the \textit{brain}). In this plot, we again observe that the maximum contribution in terms of error is from the \textit{number of estimators} hyper-parameter. Similar to the error contribution from steps and algorithms, we observe that even though, Bayesian optimization follows the trend in grid search, it is not able to accurately capture the contribution, especially for the \textit{number of estimators} hyperparameter.

\begin{figure}[ht!]
    \centering
    \includegraphics[width=0.5\textwidth]{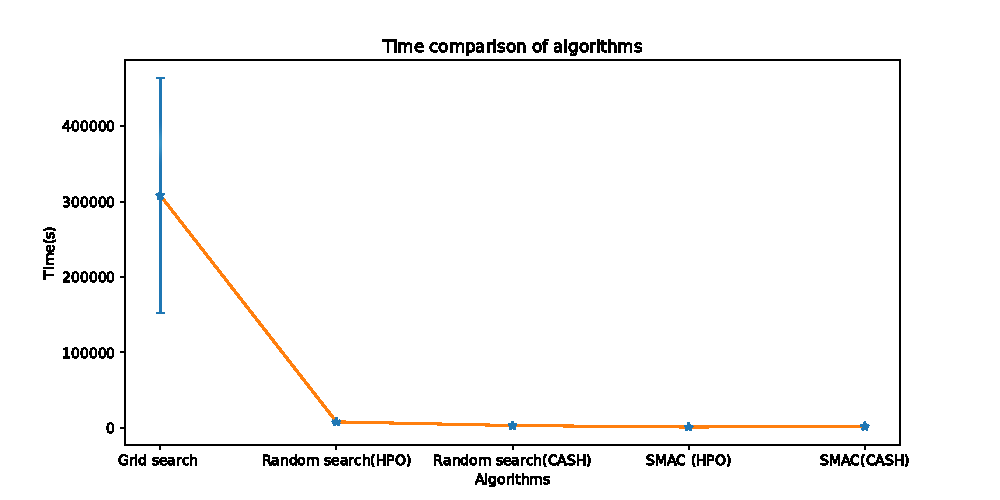}
    \caption{Comparison of computational time for running each of the 3 optimization methods in the two optimization frameworks (HPO and CASH) in section \ref{subsec_AS_HPO} averaged over the 4 datasets in Table \ref{table:datasets}. Random search and Bayesian optimization are both much more efficient than grid search in terms of computation time and maybe used to quantify the error contribution more efficiently than grid search.}
    \label{fig:time}
\end{figure}

\subsubsection{Comparison of computation time}
Fig. \ref{fig:time} shows the average computation times of each of the algorithms used in the error quantification experiments. We can observe that random search and Bayesian optimization are both efficient in terms of computational time as opposed to grid search.
Therefore, both CASH and HPO optimization frameworks using algorithms like Random search and Bayesian optimization maybe used for computation of the error contributions instead of grid search. However, as we have seen repeatedly from the plots in Figs. \ref{fig:eq_steps}, \ref{fig:eq_alg} and \ref{fig:eq_hyper}; random search captures the behavior of grid search more accurately and is more robust than grid-search. Hence, random search maybe used to quantify the contributions from steps, algorithms and hyperparameters accurately and efficiently.

\section{Application}
We describe an ongoing application of this work in a project known as the Cognitive Immersive Systems Laboratory \cite{su2017cognitive} (CISL) at Rensselaer Polytechnic Institute (RPI). The next Cognitive and Immersive Situations Room (CIR) to augment group intelligence is being built here.
\begin{figure}[ht!]
    \centering
    \includegraphics[width=0.4\textwidth]{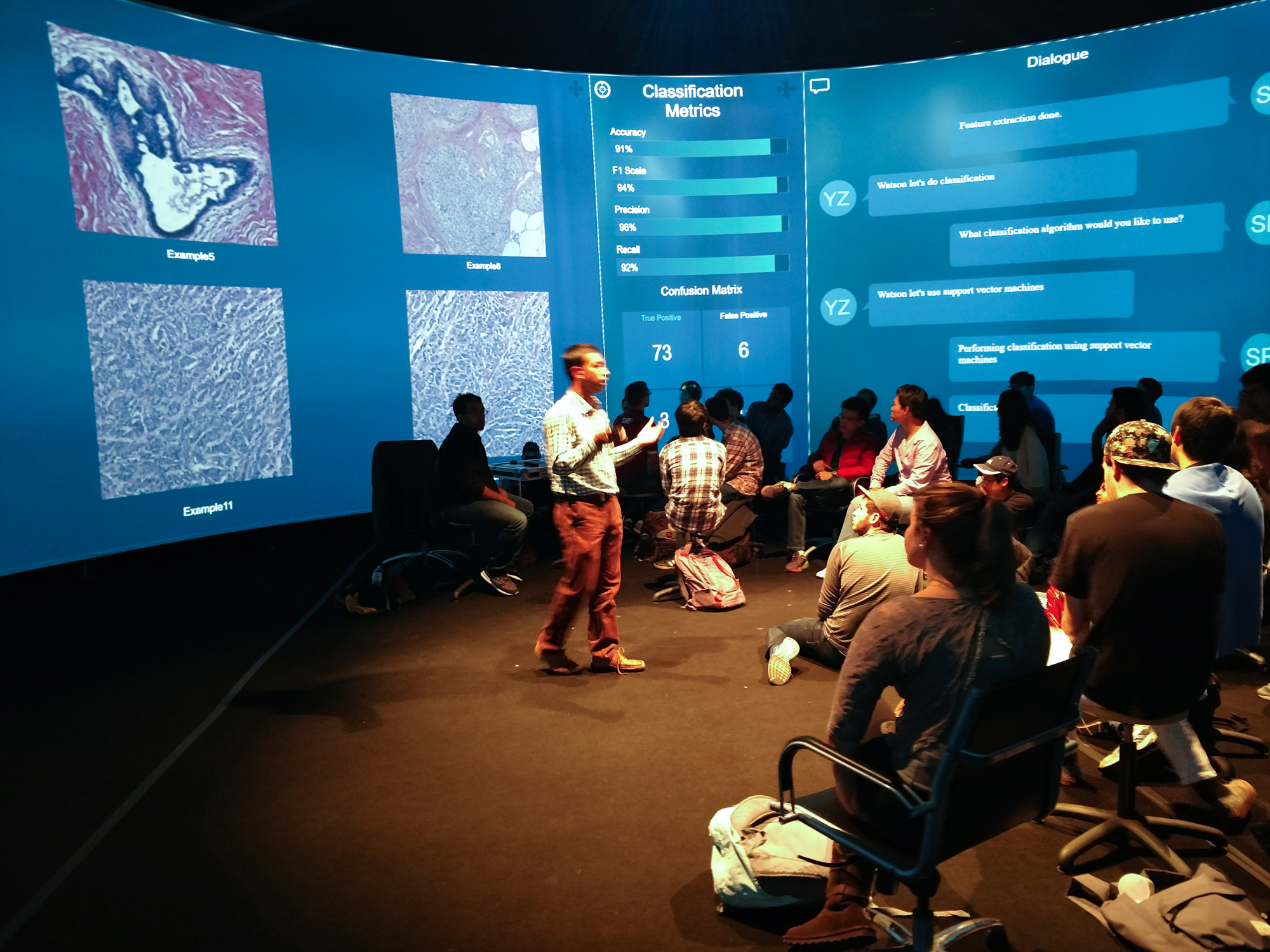}
    \caption{Example of an application of the error contribution framework. This figure shows a demonstration of the CIR using a panoramic display for the use case of breast cancer diagnosis. The front-end of the display is depicted in Fig. \ref{fig:CISL}.}
    \label{fig:CIR}
\end{figure}

A use case of CISL is to diagnose breast cancer with the help of experts consisting of medical professionals (domain experts) and computer scientists (data scientists). This group of experts come together to solve the problem with the help of the artificial intelligence in the room (CIR).

A demonstration of the CIR for breast cancer diagnosis is shown in Fig. \ref{fig:CIR}. The frontend of the CIR for this specific use-case is shown in Fig. \ref{fig:CISL}. The experts in the room look at various aspects of the problem which are depicted in the figure. This is an ideal application of the error contribution framework where the experts can look at the contributions of the various components in Fig. \ref{fig:CISL} and decide on which components to optimize and improve the image classification pipeline.

\section{Conclusion}
 The suggested approach involves understanding the sources of error contribution in data analysis pipelines. Specifically, we propose an \textit{agnostic} methodology to quantify the error contributions from different parts of an image classification pipeline, namely computational steps, algorithms and hyperparameters. This is described in Section \ref{EQ}. The results in Section \ref{eq_expts} show that the global optimization methods like random search and Bayesian optimization is able to quantify the error contributions as well as grid search. The framework of Bayesian optimization is not as accurate and robust as random search due to reasons specified in section \ref{sec4}. In general we expect optimization algorithms that have more trials in a larger region of the search space of the configurations to quantify the contributions from components in the pipeline accurately. We intend to explore the results from more hyper-parameter optimization algorithms in the future. 
The \textit{agnostic} error contribution methodology maybe used by machine learning practitioners to understand and interpret results of a specific machine learning problem on a dataset. Understanding the sources of contribution will help data scientists quickly iterate over pipelines, algorithms and hyperparameters and find the best set of configurations for solving a particular task by focussing on the important components of the pipeline found from the results. Domain experts like biologists and scientists from different disciplines can use this method to understand and interpret the error in terms of the pipeline used to solve their specific problem.
In terms of future work, this formulation could be expanded to cover more data analysis problems more algorithms. The error quantification framework maybe used by any practitioner that works with pipelines for solving a machine learning problem. The pipeline may even be extended to include processes like data cleaning and data normalization. This framework may also be included to understand and interpret deep neural networks, which are more end-to-end in nature. This maybe used for comparing the performance of candidate networks for solving the problem.

\label{sec5}
\bibliographystyle{IEEEtran}
\bibliography{sample}
\end{document}